\documentclass[11pt,a4paper]{article}
\usepackage[hyperref]{naaclhlt2018}
\usepackage{times}  
\usepackage{helvet}  
\usepackage{courier}  
\usepackage{url}  
\usepackage{graphicx}  
\usepackage{booktabs}
\usepackage{amsmath}
\usepackage{multicol}
\usepackage[inline]{enumitem}
\usepackage{algorithm,algpseudocode}

\usepackage{url}

\aclfinalcopy 
\def\aclpaperid{***} 

\title{Query Focused Abstractive Summarization: Incorporating Query Relevance, Multi-Document Coverage, and Summary Length Constraints into seq2seq Models}

\author{Tal Baumel \\
  Dept. of Computer Science  \\
  Ben-Gurion University  \\
  Beer-Sheva, Israel \\
  {\tt talbau@cs.bgu.ac.il} \\\And
  Matan Eyal \\
  Dept. of Computer Science  \\
  Ben-Gurion University  \\
  Beer-Sheva, Israel \\
  {\tt mataney@cs.bgu.ac.il} \\\And
  Michael Elhadad \\
  Dept. of Computer Science  \\
  Ben-Gurion University  \\
  Beer-Sheva, Israel \\
  {\tt elhadad@cs.bgu.ac.il} \\}

\date{}

\begin{document}
\title{Query Focused Abstractive Summarization:\\Incorporating Query Relevance, Multi-Document Coverage, and Summary Length Constraints into seq2seq Models}

\maketitle

\begin{abstract}
Query Focused Summarization (QFS) has been addressed mostly using extractive methods. Such methods, however, produce text which suffers from low coherence. 
We investigate how abstractive methods can be applied to QFS, to overcome such limitations.  Recent developments in neural-attention based sequence-to-sequence models have led to state-of-the-art results on the task of abstractive generic single document summarization. Such models are trained in an end to end method on large amounts of training data. We address three aspects to make abstractive summarization applicable to QFS: \begin{enumerate*}[label=(\itshape\alph*\upshape)]
\item since there is no training data, we incorporate query relevance into a pre-trained abstractive model;
\item since existing abstractive models are trained in a single-document setting, we design an iterated method to embed abstractive models within the multi-document requirement of QFS;
\item the abstractive models we adapt are trained to generate text of specific length (about 100 words), while we aim at generating output of a different size (about 250 words);
\end{enumerate*}  
we design a way to adapt the target size of the generated summaries to a given size ratio.  We compare our method (Relevance Sensitive Attention for QFS) to extractive baselines and with various ways to combine abstractive models on the DUC QFS datasets and demonstrate solid improvements on ROUGE performance.

\end{abstract}

\section{Introduction}
The query-focused summarization (QFS) task was first introduced in  DUC 2005 \cite{dang2005overview}. This task provides a set of queries paired with relevant document collections, each collection sharing a topic. The expected output is a short summary answering the query according to data in the documents.  Current state-of-the-art methods for the task \cite{DBLP:journals/corr/abs-0907-1814,fisher2006query,Feigenblat:2017:UQM:3077136.3080690} are extractive, {\em i.e.}, the produced summary is a set of sentences extracted from the document set. 

Extractive methods tend to produce less coherent summaries than manually crafted ones. Some examples of the weaknesses of extractive methods include unresolved anaphora, unreadable sentence ordering, and lack of cohesiveness in text. Another problem of extractive methods is the lack of ability to extract salient information from a long sentence without including less salient information included in the sentence: once the system is committed to a sentence, the full sentence will be extracted.  It has been well documented that extractive algorithms \cite{haghighi2009exploring, nallapati2017summarunner} tend to prefer longer sentences.

While most of the reasons for the weaknesses of extractive summarization methods are hard to quantify, we can illustrate the high probability of achieving incoherent text when applying extractive methods for QFS. We assume that a sentence cannot be well understood without its context if it starts with a connective phrase (which we identify by matching a closed set of connectives) or breaks a co-reference chain (sentences where a non-proper definite noun phrase or a pronoun refers to a noun phrase from a preceding sentence -- we identified co-reference chains using core-NLP \cite{lee2013deterministic}). The percent of sentences in DUC 2007 that passed the two conditions mentioned was lower than 11\%, so only 11\% of the sentences in DUC 2007 can be understood without context. 

This data on the risks on producing low coherence text is a great incentive to test abstractive summarization methods for the task of QFS. In this work, we aim at adapting abstractive single document summarization methods to handle the QFS task.  

\begin{figure} 
  \centering 
  \includegraphics[width=2.5in]{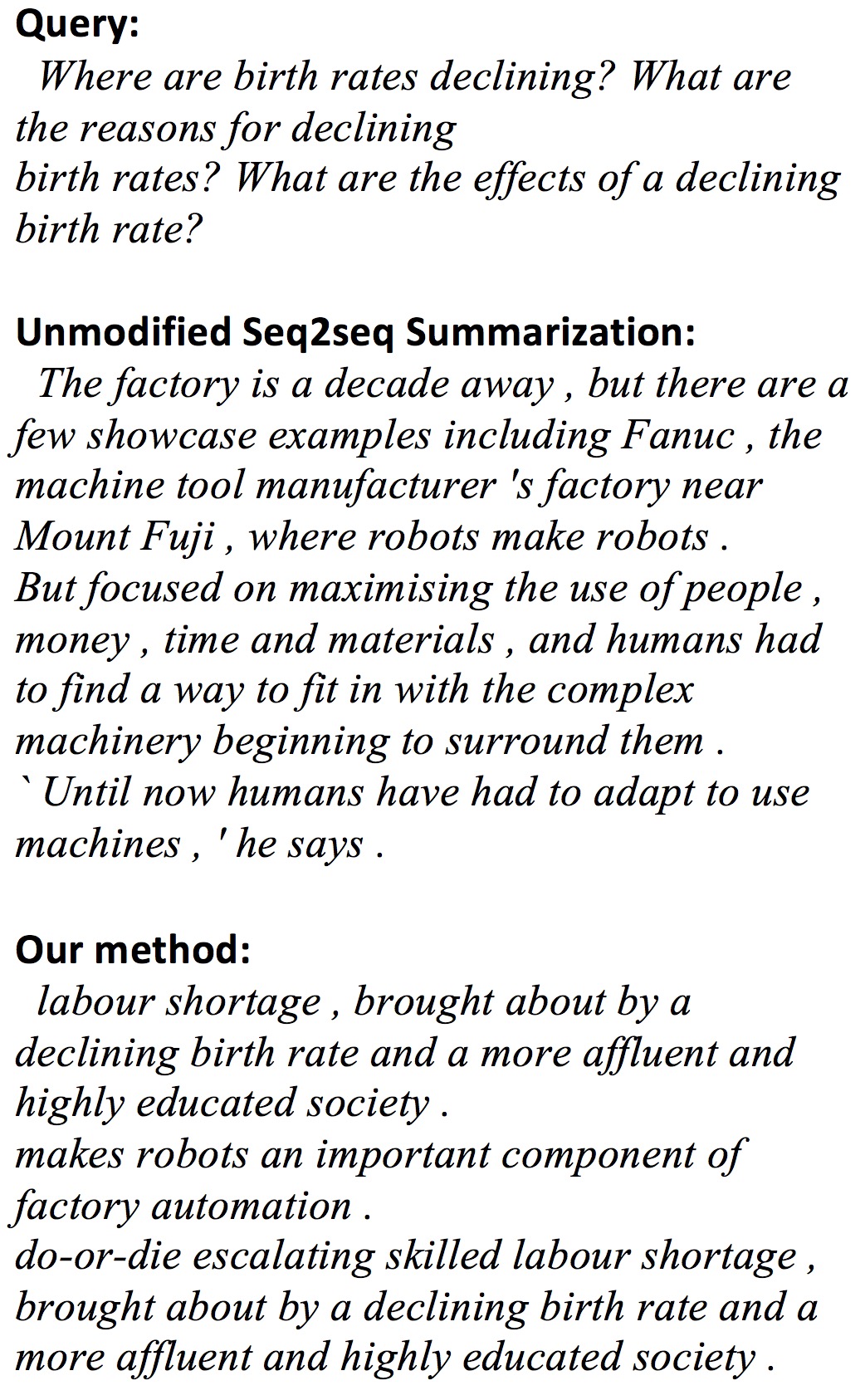} 
  \caption{Comparison of the output of the unmodified seq2seq model of See \emph{et al}. vs. our model RSA-QFS on a QFS data sample. The unmodified summary lacks coherence and is not relevant to the input query.}
  \label{fig:RSAexample} 
\end{figure}

The first obstacles we face are: 
\begin{enumerate*}[label=(\itshape\alph*\upshape)]
\item no training data is available for training end to end QFS in a way similar to what was recently done for single document generic abstractive summarization;
\item existing abstractive models cannot handle multiple documents as input and do not include an explicit query relevance criterion in their computation of salient summary content;
\item existing abstractive models have been trained to produce short summaries regardless of the information density of the input document.
\end{enumerate*}  For all these reasons, a direct application of an existing state of the art abstractive model to a QFS data sample produces inappropriate output (see Fig.~\ref{fig:RSAexample}).

We hypothesize that an existing trained abstractive model encapsulates reusable linguistic knowledge which we can leverage for the QFS task. We investigate ways to augment such a pre-trained single document abstractive model with explicit modeling of query relevance, the ability to handle multiple input documents and to adjust the length of the produced summary accordingly.  

We validate this hypothesis within the framework of a sequence to sequence (seq2seq) architecture with attention mechanism which has been adopted by most abstractive approaches.  We systematically explore the stage at which query relevance is most beneficial to the QFS abstractive process.  Further, we experiment with a method to build a summary through an iterative process of extraction / abstraction pairs: batches of relevant content from multiple documents are identified, then abstracted into a sequence of coherent segments of text. 

We compare our system both with top extractive methods and with various combinations of a pre-trained abstractive model with relevance matching and multiple document input.  We evaluate the proposed model, called Relevance Sensitive Abstractive QFS (RSA-QFS), on the traditional DUC datasets.  Our experiments demonstrate the potential of abstractive QFS models with solid ROUGE gains over those baselines.

\section{Previous Work}

\subsection{Extractive Methods}

Current state-of-the-art methods on the task of QFS on the DUC dataset could be categorized into unsupervised methods and small scale supervised methods:

\paragraph*{Unsupervised methods} search for a set sentences that optimizes a gain function. The \emph{ Cross Entropy Summarizer} (CES) \cite{Feigenblat:2017:UQM:3077136.3080690} optimizes relevance under length constraint.  This method achieves current state-of-the-art ROUGE scores on DUC 2005-7 datasets.

\paragraph*{Small scale supervised methods} use small datasets (usually previous DUC datasets) to learn a representation of the dataset, and using this representation, optimize a gain function. A recent example of this approach is  \emph{DocRebuild} \cite{ma2016unsupervised}, which trains a neural network to find a set of sentences that minimize that original document reconstruction error. The method uses DUC 2006-7 to learn word representations, and obtains results slightly lower than CES.

All of the extractive methods suffer from the coherence problems mentioned above.

\subsection{Sequence-to-Sequence Models for Abstractive Summarization}

Abstractive methods in generation have emerged as practical tools since 2015.  At this point, the most successful attempts at abstractive summarization are on the task of generic single document summarization \cite{rush2015neural,nallapati2016abstractive,DBLP:journals/corr/SeeLM17,DBLP:journals/corr/PaulusXS17} and are based on the sequence-to-sequence (seq2seq) approach with attention mechanism \cite{DBLP:journals/corr/BahdanauCB14}. These models include the following components:

\paragraph{Encoder:} a neural network transforms a list of words into a list of dense vector representations. These dense representations aim to capture both the word and its context. Encoders are most commonly implemented using a word embedding layer followed by a Recurrent Neural Network (RNN), \emph{i.e.}, a Long Short Term Memory (LSTM) component \cite{hochreiter1997long} or Gated Recurrent Units (GRU) \cite{chung2014empirical}).

\paragraph{Decoder:} a neural network generates the next word in the summary conditioned on the representation of the prefix of the generated text and a dense context vector representing the input sequence.  The decoder is commonly implemented by an RNN, a fully connected layer with the dimension of the output matching the size of the vocabulary, and a softmax layer that turns a vector into a distribution over the vocabulary.

\paragraph{Attention mechanism:} a neural network determines the importance of each encoded word at each decoding step, and maps the variable length list of encoded words representations into a fixed-size context representation.  The attention mechanism is commonly implemented using multiple levels of fully connected layers to calculate the unnormalized attention weight of each word in the input and a softmax layer to normalize these weights. 

\paragraph*{}The training of such models for abstractive single document summarization has been made possible by the availability of large scale datasets such as GIGAWORD \cite{graff2003english}, the New-York Times dataset \cite{sandhaus2008new} and CNN/Daily~News \cite{nips15_hermann}, which contain pairs of source text / short text examples. 
For example, the CNN/Daily~Mail Corpus was automatically curated by matching articles to the summary created by the site editor.  The dataset includes 90K documents from CNN and 196K documents from the Daily~Mail. The average size of an abstract in the corpus is $\sim$100 words, and the size of input documents is about 1,000 words. In contrast, the average abstract length in the DUC QFS dataset is $\sim$250 words.

No such large scale dataset is currently available for the QFS task under the DUC settings. We hypothesize that models trained on such datasets capture the linguistic capability to combine small windows of coherent sentences into concise paraphrases.  Accordingly, our objective is to adapt such a pre-trained, generic abstractive summarization architecture to the more complex task of QFS.

Recent work in abstractive QFS summarization \cite{nema2017diversity} attempt to solve the issue of missing training data by introducing a new dataset for abstractive QFS based on \emph{Debatepedia}. The dataset introduced is, however, very different from the DUC QFS datasets since the summaries presented are debate key points that are not more than a single short sentence with on average 10 words per summary vs. 250 words in the DUC data; the input texts are also short snippets of text with an average of 60 words vs. DUC that can reach more than 1,000 words. Because of the distinct size differences between the DUC and \emph{Debatepedia} datasets, we cannot compare the methods directly.

In this work, we focus on adapting a specific architecture: the pointer-generator with coverage mechanism network of \cite{DBLP:journals/corr/SeeLM17}, to the QFS task. This model achieves state-of-the-art ROUGE \cite{lin2004rouge} and readability scores on the single document generic abstractive summarization task. 
Although the pointer-generator with coverage mechanism network includes significant modifications (pointer-network and coverage mechanisms), it still adheres to the general encoder-decoder-attention architecture. We thus present our proposed modification in the simplified context of the generic architecture, as the handling of relevance is orthogonal to the processing of rare words using switch-generator and the coverage mechanism ability to avoid redundancy.  Our experiments are using the full network.

\section{Query Relevance}

\begin{figure}
  \centering 
  \includegraphics[width=2in]{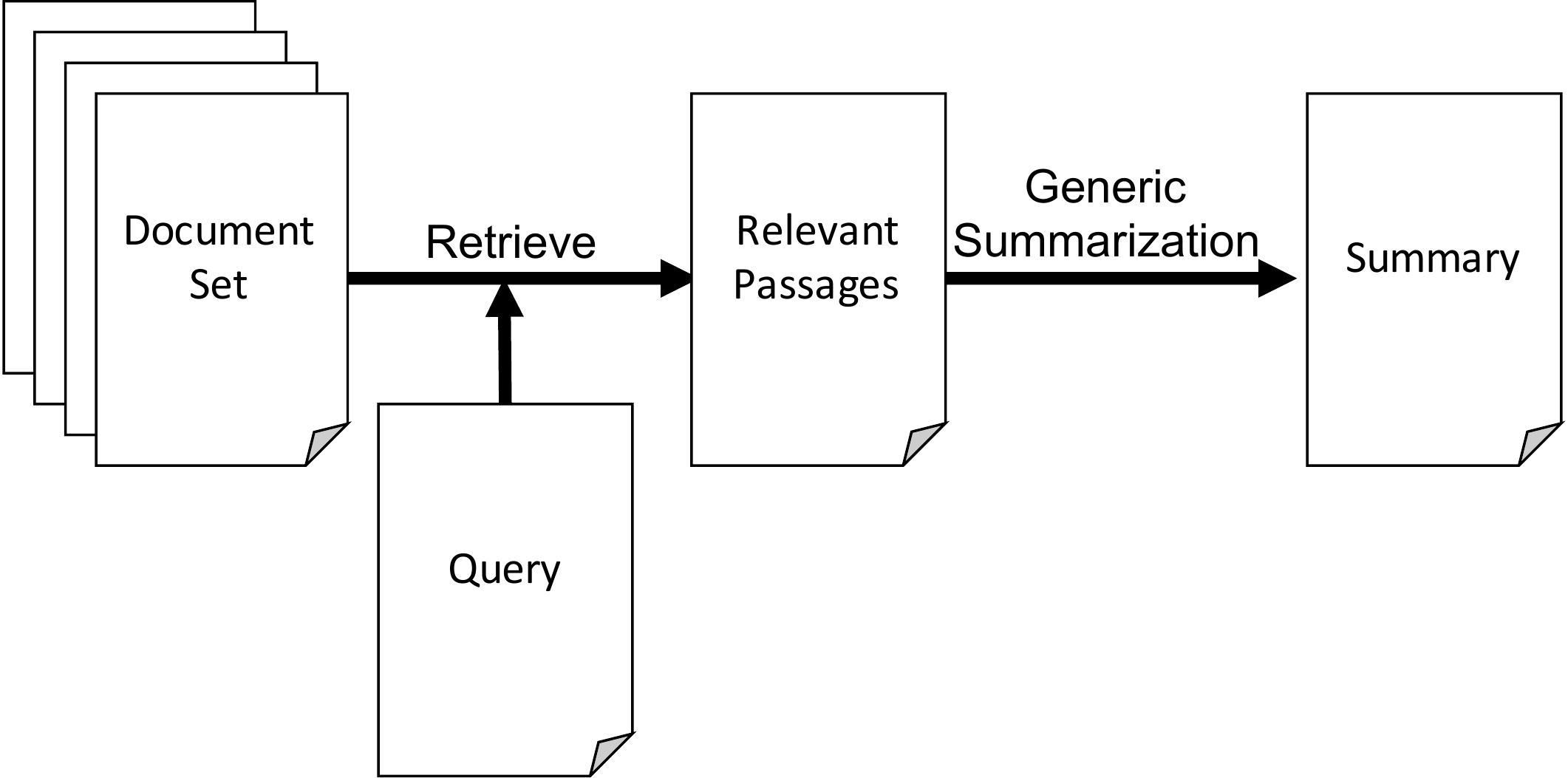} 
  \caption{Two stage query focused summarization scheme.}\label{fig:RSAtwostage} 
\end{figure}

We adopt the approach to QFS formulated in \cite{baumel2016topic}: the QFS task is split into two stages, a \emph{relevance model} determines the extent to which passages in the source documents are relevant to the input query; and a generic summarization method is applied to combine the relevant passages into a coherent summary. The relevance model identifies redundant, un-ordered passages using Information Retrieval methods; 
whereas the summarization model selects the most salient content, removes redundancy and organizes the target summary.  This schematic approach is illustrated in Figure~\ref{fig:RSAtwostage}). This method achieves good ROUGE results when using simple extractive summarization methods such as KL-sum \cite{haghighi2009exploring} when the relevance model is of high quality.  

Accordingly, in order to adapt abstractive methods to QFS, the first baseline we consider consists of filtering the input documents according to relevance and then pass the filtered relevant passages to an abstractive model.  We hypothesize that this approach will not adapt well for abstractive methods because the input that is generated by the filtering process is quite different from the type of documents on which the abstractive model was trained: it is not a well structured coherent document. Abstractive models rely critically on the sequential structure of the input to take decision at generation time. Our method aims at preserving the document structure while infusing relevance into the abstractive model during decoding.

In this paper, we consider very simple relevance models and do not attempt to optimize them -- we compare relevance measures based on unigram overlap between query and sentences, and TF*IDF and Word2vec encodings with cosine distance between the query and sentences.  To get an upper bound on the impact a good relevance model can have, we also consider an Oracle relevance model, where we compare sentences with the gold summaries using the word count cosine measure.  Our focus is to assess whether the mechanism we propose in order to combine relevance and abstractive capabilities is capable of producing fluent and relevant summaries given a good relevance model.

\section{Methods}

\subsection{Incorporating Relevance in Seq2Seq with Attention Models}

As discussed above, the lack of a large scale dataset similar to QFS task presented in DUC 2005-7 prevents us from attempting an end-to-end solution that learns to generate a relevant summary using a the documents and the query as input. In order to overcome this obstacle, we split the problem in two tasks - a relevance model and an abstractive model that takes relevance into account.  Relevance can be introduced into an existing seq2seq with attention model in different ways: 

\begin{enumerate*}[label=(\itshape\alph*\upshape)]
\item Filter the input to include only sentences with high relevance score and pass the filtered input to the model at generation time; we test this method as a baseline. 
\item Inject the relevance score into the pre-trained model.
\end{enumerate*}

Given a document and a query, we calculate the relevance of each sentence to the query (as a pre-processing step) and use this relevance as an additional input to the network.  The relevance model predicts the relevance of a sentence given the query.  We project the relevance score of sentences to all the words in the sentence to obtain a word-level relevance score.  

\begin{figure}
  \centering 
  \includegraphics[width=3in]{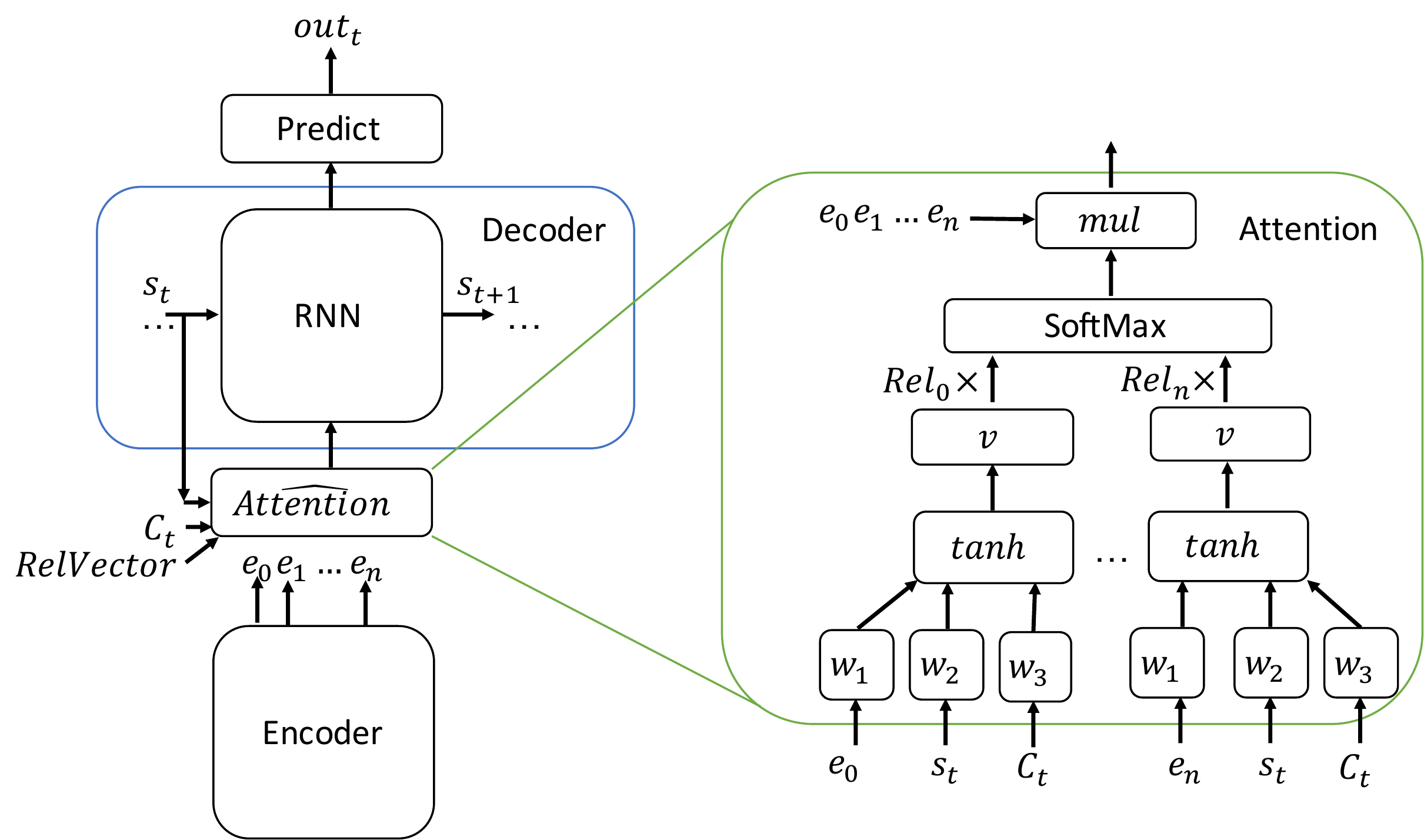} 
  \caption{Illustration of the RSA-QFS architecture: $RelVector$ is a vector of the same length as the input ($n$) where the $i$th element is the relevance score of the $i$th input word. $RelVector$ is calculated in advance and is part of the input.}\label{fig:RSAmod} 
\end{figure}

At each decoding step in the abstractive seq2seq model, we multiply each (unnormalized) attention score of each word calculated by the model by the pre-computed relevance score, as illustrated in Fig.~\ref{fig:RSAmod}. In the unmodified seq2seq model we adapted \cite{DBLP:journals/corr/SeeLM17}, the (unnormalized) attention of word $i$ at step $t$ is calculated by: 

\begin{align*}
e^t_i=v^T\tanh(W_hh_i+W_sS_t+b_{attn})
\end{align*}

where $v, W_h, W_s$ and $b_{attn}$ are trained parameters, $h_i$ is the encoder output for word $i$, and $S_t$ is the decoder state at step $t-1$. The attention scores are later normalized using a softmax function. In our model, we multiply each word by its relevance score before normalization: 
\begin{align*}
	\hat{e}^t_i=Rel_i \times e^t_i 
\end{align*}
where $Rel_i$ is the relevance score of $w_i$ which combines sentence relevance and lexical relevance, as predicted using the relevance ranking model (all words in the same sentence are given the same relevance score).  We discuss below the range of relevance models with which we experimented and how the relevance scores are calibrated in the model.

In this scheme, the adapted model is able to ignore irrelevant sentences at generation time, while still benefiting from their context information at encoding time.  This is in contrast to the Filtering baseline, where the encoder is not fed low-relevance sentences at all.   We hypothesize that in our proposed scheme, the encoder will produce a better representation of the input documents than in the Filtered baseline because it is used in the same regime in which it was trained.

It is important to note that we do not re-train any of the model, the original parameters of the baseline encoder-decoder-attention model are re-used unchanged.  

\subsection{Calibrating the Relevance Score}
\begin{figure}
  \centering 
  \includegraphics[width=3in]{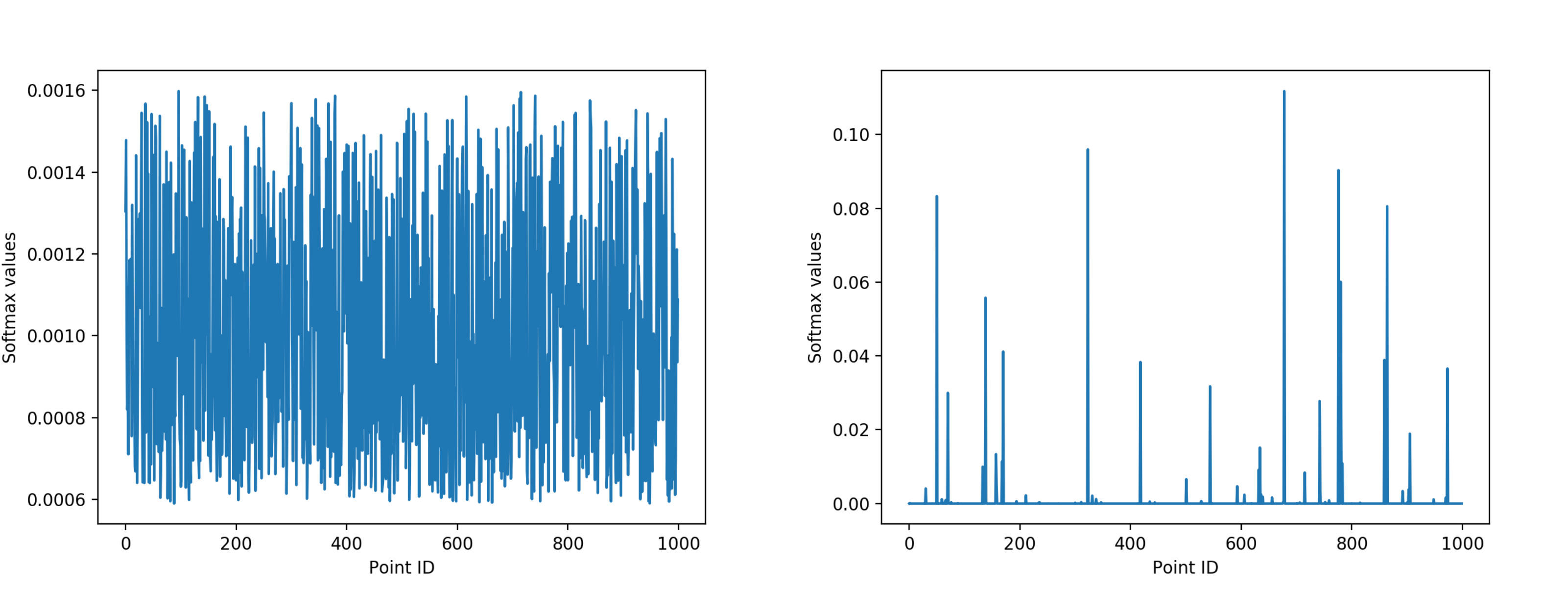} 
  \caption{A demonstration of the scale sensitivity of the softmax function. Both figures illustrate a softmax operation over 1,000 samples from a uniform distribution; left is sampled from the range 0--1 and the right from 0--100.}\label{fig:RSAsm} 
\end{figure}

Unlike other normalization methods, the softmax function is very sensitive to the scale of the input values: when the scale of the input is lower, the variance of the softmax output is similarly low (see Figure~\ref{fig:RSAsm}). When the variance of the softmax output is low, there is no single word that receives most of the normalized attention, and the model is unable to ``focus'' on a single word. Since most attention models use softmax to normalize the attention weights it is important to keep their scale when multiplying them by the relevance scores to keep well calibrated attention scores.

To address this issue, we multiplied the cosine similarity scores by 10 in order to increase the scale from 0--1 to 0--10 before applying softmax normalization. This scale modification had a significant impact on the reported ROUGE performance.

\subsection{Adapting Abstractive Models to Multi-Document Summarization with Long Output}

Summarization datasets such as the Daily-Mail/CNN include single-document input and short summary (about 100 words where DUC requires 250 words).  We need to adapt the pre-trained abstractive model to handle the multi-document scenario and produce longer output.

One possible solution is to use an extractive summarization method to generate the input and apply an abstractive method over it. While this method may handle multiple documents as input, it suffers from two problems: 
it can only decrease recall since it is unlikely that the abstractive method can introduce relevant information not included in the input and it will suffer from the abstractive model bias for short output -- we cannot directly encourage the abstractive model to generate longer text to cover more content.

Instead, we use the following simple eager algorithm to produce summaries from multiple documents and control the length of the output. We first sort the input documents by overall TF-IDF cosine similarity to the query, then iteratively summarize the documents till the budget of 250 words is achieved. 
To avoid redundancy, we filter out generated sentences from the generated summary when more than half of their words are already included in the current summary. 

\begin{algorithm} 
\begin{small}
  \begin{algorithmic}
  \State $documents \Leftarrow sort\_by\_relevance(documents)$
  \State $output\_summary \Leftarrow new\_summary()$
  \For{$document \in documents$}
  	\State $summary \Leftarrow RSA\_word\_count(document)$
  	\For{$sentence \in summary$}
    	\If{$len(output\_summary + sent) > budget$}
        	\State $return\:output\_summary$
        \EndIf
        \If{$is\_novel(output\_summary, sentence)$}
        	\State $output\_summary += sentence$
        \EndIf
  	\EndFor
  \EndFor
  \end{algorithmic}
 \end{small}
  \caption{Iterative Version}
      \label{RSAalgo}
\end{algorithm}

This algorithm ignores document structure and topic progression and uses a simplistic model of redundancy.  We leave for future work the comparison of this baseline algorithm with more sophisticated models of content redundancy and discourse.

\section{Experiments}
The goals of the experiments are: 
\begin {enumerate*} [label=(\itshape\alph*\upshape)]
\item to compare RSA-QFS with the baseline where the input documents are filtered according to relevance;
\item to test whether the method to incorporate relevance within the attention mechanism on a single document input produces readable and relevant output;  
\item to measure the impact of the quality of different relevance models on the output of RSA-QFS on a single document input;
and \item to evaluate the iterative version of RSA-QFS vs. a state-of-the-art extractive QFS method (CES).
\end{enumerate*}

\subsection{Evaluation}

We tested the various scenarios using the QFS track data from the DUC 2005, 2006 and 2007 datasets \cite{dang2005overview,hoa2006overview}.  We also compared RSA-QFS on the \emph{Debatepedia} dataset despite the differences in sizes discussed above. We use ROUGE metrics for all performance comparisons.

We evaluate separately the incorporation of the relevance model with a pre-trained abstractive model on a single document as an ablation study, and we test the iterative algorithm to handle multiple input documents in a second round of experiments.

In the first round of experiments, we compare various abstractive baselines on the longest input document from the QFS topic set (we also experimented with the most relevant document but obtained lower ROUGE performance). For such comparisons, we use ROUGE-1, ROUGE-2 and ROUGE-L metrics (ROUGE-L measures recall on the longest common substrings).  When comparing RSA-QFS to the extractive method, we use ROUGE-1, ROUGE-2 and ROUGE-SU4 which are most usually reported for extractive method performance.

The ROUGE values obtained in the single-document ablation study are expected to be much lower than competitive QFS results for two main reasons: (1) we use as reference the DUC reference summaries with no modifications. These reference summaries were created manually to cover the full topic set; in contrast, in the ablation study we read only a single document, and the best summary we can generate will lack coverage; (2) the pointer-generator abstractive model was trained to generate a $\sim$100 words summary, while DUC datasets summaries contain $\sim$250 words.  Still the reported ROUGE performance indicates trends to detect whether the generated short summaries manage to capture relevance.

\subsection{Abstractive Baselines}

We compare RSA-QFS with the following baselines:

\paragraph{BlackBox:} We run the document through the pointer-generator abstractive model without any modifications. This weak baseline indicates whether our method improve QFS performance vs. an abstractive method that completely ignores the query.   

\paragraph{Filtered:} We filtered half of the document sentences by selecting the ones with the highest relevance score. We maintained the original ordering of the sentences. We then used the filtered document as an input to the pointer-generator abstractive model. The relevance score we used was the count of shared words betwen the query and the sentence (see below the list of other relevance models we tested - this model of relevance provided the best results on the filtered baseline).

\paragraph{Relevance Sensitive Attention (RSA-QFS):} This method is the main contribution of this work. 

We tested the method using the following relevance score functions:
\paragraph{Word Count} is a simple count of word overlap between the query and a given sentence.

\paragraph{RSA-TFIDF:} We generated a \textbf{TF-IDF} representation of the entire document set for each topic and aggregated the sentence scores using cosine similarity between the query and the sentence TF-IDF vectors (\textbf{RSA TFIDF}).

\paragraph{RSA word2vec:} We use a \textbf{word2vec} model \cite{mikolov2013distributed}, pre-trained on the Google News dataset. Relevance is measured as the cosine similarity between the summed representation vector of each word in the query and in the sentence. Words that did not appear in the pre-trained word2vec model vocabulary were ignored.

Results are given in Table~\ref{RSAresultsSingle}.  
As expected, the ``Blackbox'' method which ignores the query completely performs poorly.  
More surprisingly, we observe that the Filtered model (where we filter the input document according to the word-count relevance model and then apply the abstractive model) does not behave any better than the blackbox unmodified model.  In contrast, RSA-QFS improve significantly (all improvements are significant within 5\% except DUC-2005 ROUGE-2) over the Filtered pipeline - while processing exactly the same input material as the Filtered method.  This indicates that the way we incorporate relevance within the Attention mechanism is more effective than directly adjusting the input representation.

The word count relevance model achieves the highest ROUGE scores when compared with other relevance models. On all the datasets, it outperforms the filtered baseline by a large amount. The word2vec-based method is close and consistently within confidence interval of the word count method. We speculate that the fact that out of vocabulary words are ignored, and the fact that DUC queries tend to be verbose and do not need much expansion explain the fact that word2vec does not improve on the Word count model. The TFIDF-based method performed poorly. We presume this is due to the fact that the ROUGE settings\footnote{PyRouge default settings as used in the pointer-generator evaluation \cite{DBLP:journals/corr/SeeLM17}} did not eliminate stop words and frequent words for the evaluation.

\begin{table*}[t]
\centering
 \begin{small}
    \begin{tabular}{l|ccc|ccc|ccc}
    \toprule
     & \multicolumn{3}{c}{DUC2005}& \multicolumn{3}{c}{DUC2006}& \multicolumn{3}{c}{DUC2007} \\ 
    Single Document & 1 & 2 &  L & 1 & 2 &  L & 1 & 2 &  L \\
    \midrule
    BlackBox & 12.11 & 1.38 & 11.28 & 14.76 & 2.29 & 13.35 & 15.33 & 2.68 & 14.04\\
    Filtered  & 12.09 & 1.33 & 11.20 & 14.71 & 2.25 & 13.48 & 16.23 & 2.89 & 14.76\\
    RSA Word Count  & \textbf{12.65} & \textbf{1.61} & \textbf{11.79} & \textbf{16.34} & \textbf{2.56} & \textbf{14.69} & ֿ\textbf{17.80} & \textbf{3.45} & \textbf{16.38} \\
    RSA TFIDF   & 11.84 & 1.57 & 11.00 & 13.93 & 2.12 & 12.90 & 14.71 & 2.61 & 13.50\\
    RSA word2vec  & 12.44 & 1.39 & 11.52 & 15.80 & 2.43 & 14.46 & 17.55 & 3.21 & 15.90\\
    \bottomrule 
    \end{tabular}
  \end{small}
 \caption{Incorporating Relevance on a Single (Longest) Document Input}
      \label{RSAresultsSingle}
\end{table*}

\begin{table*}[t]
\centering
 \begin{small}
    \begin{tabular}{l|ccc|ccc|ccc}
    \toprule
     & \multicolumn{3}{c}{DUC2005}& \multicolumn{3}{c}{DUC2006}& \multicolumn{3}{c}{DUC2007} \\ 
    Multi-Document & 1 & 2 &  SU4 & 1 & 2 &  SU4 & 1 & 2 &  SU4 \\
    \midrule
    CES & \textbf{40.33} & \textbf{7.94} & 13.89 & \textbf{43.00} & \textbf{9.69} & 15.63 & \textbf{45.43} & \textbf{12.02} & 17.50\\
    Iterative RSA Word Count& 39.82 & 6.98 & \textbf{15.73} & 42.89 & 8.73 & \textbf{17.75} & 43.92 & 10.13 &  \textbf{18.54}\\
    Iterative RSA Oracle & \emph{43.48} & \emph{8.75} & \emph{17.94} & \emph{46.64} & \emph{10.96} & \emph{20.34} & \emph{47.91} & \emph{12.77} & \emph{21.37}\\
    \bottomrule 
    \end{tabular}
  \end{small}
 \caption{Iterative RSA-QFS vs. Extractive Methods}
      \label{RSAresultsMulti}
\end{table*}

\subsection{Extractive Baselines}

In this part of the experiments, we compare the RSA-QFS method extended with the iterative algorithm to consume multiple documents and a query under the exact DUC conditions and produce summaries comparable to existing extractive methods.  We compare with CES, the current state of the art extractive algorithm on QFS.  Results are in Table~\ref{RSAresultsMulti}.  

We compare two relevance models with RSA-QFS: the word-count model which we identified as the best performing one in the ablation study, and the Oracle model. In the Oracle model, we compute the relevance of an input sentence by comparing it to the reference models instead of with the query.  This gives us a theoretical upper bound on the potential benefit of more sophisticated retrieval ranking methods.

We observe that RSA-QFS is competitive with state-of-the-art extractive methods and outperforms them in the SU4 metric. The oracle baseline shows that a more sophisticated relevance method has the potential to improve performance by a significant amount and way above the current extractive top model.

\subsection{Evaluation Using the Debatepedia Dataset}
We used the Debatepedia QFS dataset \cite{nema2017diversity} to evaluate our method vs. the LSTM based diversity attention trained end to end on the Debatepedia dataset ($SD_2$) model. We compare with the ROUGE recall results provided in the original paper. While the result in Table \ref{RSAdebatepedia} may suggest our method outperforms the model that is trained on the actual dataset, it must be noted that our model yielded summaries ten times longer than required and achieved very low ROUGE precision. We did not compare precision score since it was not provided in the original research.  This comparison indicates the datasets are not directly comparable, but that even on a completely different domain, the abstractive capability encapsulated in the model provides readable and realistic summaries.

\begin{table}
\centering
 \begin{small}
    \begin{tabular}{l|ccc}
    
    \toprule
    & \multicolumn{3}{c}{Debatepedia}\\ 
    Recall-ROUGE & 1 & 2 &  L \\
    \midrule
    $SD_2$ & 41.26 & \textbf{18.75}  & 40.43 \\
    RSA Word Count  & \textbf{53.09} & 16.10 & \textbf{46.18} \\
    \bottomrule 
    \end{tabular}
  \end{small}
 \caption{Results for Debatepedia QFS dataset}
      \label{RSAdebatepedia}
\end{table}

\section{Analysis}

\subsection{Output Abstractiveness}
In order to test if our model is truly abstractive, instead of simply copying relevant fragments verbatim from the input documents, we counted the amount of sentences from the summary generated by our model (using word count similarity function) which are substrings of the original text. We found that on average only about 33\% of the sentences were copied from the original document and that the average word edit-distance between each generated sentence and the most similar sentence is about 39 edits (tested on DUC 2007 summarized by the iterative RSA word count method). 

We observed that the generated sentences, while significantly different from the source sentences do not introduce many new content words.  Almost all generated words are  present in the source documents.
While these two measures indicate a good level of ``abstractiveness'', it remains a challenge to measure abstractiveness in an interpretable and quantitative manner.  Cursory reading of the generated summaries still ``feels'' very literal.  
 
We assessed readability by reading the summaries generated by the best performing methods, the \textbf{RSA Word Count} (an example can be seen in Fig.~\ref{fig:RSAexample}) and the \textbf{Oracle Based Iterative Method}. We found that the summaries produced by the single document variant maintained the readability of the unmodified model. We did notice that the coverage mechanism was affected due to our modification and some sentences were repeated in the summaries our model produced (compared to the original abstractive model). The iterative version did not suffer from repeated sentences, since they are dismissed by the algorithm, but did suffer from lack of coherence between sentences, indicating a better discourse model is required than the simple eager iterative model we used.  Improved coherence also requires better evaluation metrics than the ROUGE metrics we have used.

All the produced summaries for all the methods and the code required to produce them are available at \url{https://github.com/talbaumel/RSAsummarization}.

\section{Conclusion} 

In this work, we present RSA-QFS, a novel method for incorporating relevance into a neural seq2seq models with attention mechanism for abstractive summarization to the QFS task, without additional training. \textbf{RSA-QFS} significantly improves ROUGE scores for the QFS task when compared to both unmodified models and a two steps filtered QFS scheme, while preserving readability of the output summary. The method can be used with various relevance score functions. We compared the method with state-of-the-art extractive methods and showed it produces competitive ROUGE scores for the QFS task even with very simple relevance models and a simple iterative model to account for multiple input documents. When using an ideal Oracle relevance model, our method achieves very high ROUGE results compared to extractive methods. 

This study frames future work on multi-document abstractive summarization: we need to design quantitative measures of abstractiveness (how much re-formulation is involved in producing a summary given the input documents) and of summary coherence to overcome the known limitations of ROUGE evaluation when applied to non-extractive methods.  We also find that relevance models remain a key aspect of summarization and the gap between Oracle and practical relevance models indicates there is potential for much improvement on these models.

\bibliography{naaclhlt2018}
\bibliographystyle{acl_natbib}

\end{document}